\documentclass[pdflatex,sn-mathphys-num]{sn-jnl}
\usepackage{graphicx}
\usepackage{amsmath,amssymb,amsfonts}
\usepackage{booktabs,array}
\usepackage{algorithm,algorithmicx,algpseudocode}
\usepackage{microtype}
\usepackage{placeins}
\begin{document}
\title[TEM: Temporal State Tracking Memory]{TEM: Temporal State Tracking Memory for Large Language Model Agents}
\author*[1]{\fnm{Pengfeng} \sur{Song}}\email{sungkites@163.com}
\author[1]{\fnm{Yunkai} \sur{Zhuang}}
\author[1]{\fnm{Liye} \sur{Shi}}
\affil[1]{\orgdiv{School of Information and Intelligent Engineering},\orgname{Zhejiang Wanli University},\orgaddress{\city{Ningbo},\postcode{315100},\country{China}}}

\abstract{An agent's memory must remain useful as information changes across conversations. A new fact may replace an earlier value without making that value irrelevant to a later question. We present Temporal State Tracking Memory (TEM), a training-free framework for tracking these changes. TEM stores facts as versioned records linked to their sources. It retrieves relevant versions, places them in source order and labels historical records before passing the evidence to the answer model. Event memory and original dialogue turns provide details missing from structured facts. We evaluated TEM on 420 questions drawn from 600 sessions involving 60 personas. Across four open-weight models, key-entity hit rates reached 75.71--97.62\% overall and 96.67--100.00\% on the historical-residence subset. Adding historical labels to otherwise identical, source-ordered facts increased overall hit rates by 17.14--55.95 percentage points. The evidence was compact: 94--118 tokens per answer, or 48.75--56.73\% less than session-aware retrieval-augmented generation (RAG). A further evaluation on 1,540 LoCoMo questions tested hybrid memory under a fixed evidence budget. TEM exceeded session-aware RAG with both backbones; conservative score differences were 7.92 and 10.58 percentage points. The results support preserving fact versions and exposing their temporal status to the generator when answering questions about repeatedly updated information.}


\maketitle

\section{Introduction}

A long-term agent accumulates records that may no longer describe its user. Residence, occupation, employer and preferences can change several times across sessions \cite{ref1,ref2,ref3}. Earlier records still matter when a question refers to the past. The agent therefore needs to know which version applies to the question, including whether the latest record is relevant at all.

Suppose a user moves from Shanghai to Chengdu and then to Suzhou. Suzhou answers ``Where does the user live now?'' The question ``Where did the user live before moving to Chengdu?'' requires Shanghai. A memory that keeps only the latest address cannot answer both. Keeping all three addresses makes the historical answer available, but the model still has to identify the right one.

Persistent memory, action feedback and experience-based reflection have extended agents' use of past interactions \cite{ref4,ref5,ref6}. Work on hierarchical storage, memory updates, associative retrieval and temporal graphs has also improved access to this information \cite{ref7,ref8,ref9,ref10,ref11,ref12,ref13,ref14}. Repeated updates pose a particular problem: several retrieved values may be equally relevant to a question yet refer to different states. We examine how the memory can convey these version relationships to the answering model.

TEM records cross-session changes as fact versions without retraining the backbone. New facts are appended, and the status of earlier records is updated from their source order. At query time, relevance scores select the evidence. TEM then orders the selected facts by source and labels historical versions. This separates evidence selection from the presentation of temporal information. Event memory and original turns supply details that the fact records do not capture.

Our contribution is a history-preserving update rule that assigns state over complete fact-version groups. The evidence construction procedure carries this status into retrieval output, so an older record stays historical when the latest version is not retrieved. We evaluate both components across multiple backbones and examine hybrid memory on public multi-session questions.

\section{Related Work}

\subsection{Long-Term Agent Memory and Structured Organisation}

External memory lets an agent read and write information outside its model parameters. Hierarchical management in MemGPT and the historical storage mechanisms studied in MemoryBank and long-term memory augmentation provide ways to retain information beyond the immediate context \cite{ref7,ref8,ref9}. A-MEM, Mem0 and MemoryOS extend this work through memory associations, updates and hierarchical organisation \cite{ref10,ref11,ref12}. Temporal information enters the memory itself in Zep's knowledge graph \cite{ref13}, making it particularly relevant to state tracking. TEM examines how version status is carried from stored records into answer evidence. Memory processing cost is another concern; LightMem addresses it through staged operations \cite{ref14}.

Graph-based retrieval connects information spread across records. Associative retrieval is central to HippoRAG and its successor, while GraphRAG uses graph organisation for query-focused summarisation \cite{ref15,ref16,ref17}. Beyond these retrieval structures, MemoRAG uses global memory to guide evidence access, and MemOS considers knowledge access across the memory lifecycle \cite{ref18,ref19}.

KG-RAG combines knowledge-graph retrieval and question decomposition for multi-hop question answering without additional training \cite{ref20}. CTRN addresses complex time constraints through implicit temporal and relational representations \cite{ref21}. MGIT models temporal dependencies at multiple granularities \cite{ref22}. These graph-based question-answering methods are relevant to TEM's use of structured evidence. Our focus is the management of fact versions as conversations update a state, including how the resulting status information reaches the generator.

\subsection{Long-Term Memory Evaluation and Retrieval Augmentation}

LongMemEval and LoCoMo evaluate cross-session information use, fact updates and temporal reasoning \cite{ref23,ref24}. LongBench covers a broader set of long-context tasks \cite{ref25}. Lost in the Middle shows that the position of relevant information can affect model performance \cite{ref26}. For state tracking, finding a relevant passage does not settle which version answers the question. We study this choice in controlled update tasks and evaluate hybrid retrieval on public multi-session questions.

RAG and dense retrieval provide the basis for accessing external evidence \cite{ref27,ref28}. Self-RAG evaluates evidence, RAPTOR organises it hierarchically, and interleaved retrieval supports chain-of-thought reasoning \cite{ref29,ref30,ref31}. LLMLingua, LongLLMLingua and RECOMP reduce the input passed to the generator \cite{ref32,ref33,ref34}. TEM retains version order and state attributes alongside retrieved facts, allowing the generator to interpret the selected evidence in its temporal context.

\section{Temporal State Tracking Memory}

TEM stores each fact with its provenance and places successive values in a version group. Source order determines the status of each version. Retrieval then selects the facts to include in the answer evidence. Figure 1 gives the overall framework; Fig. 2 and Algorithm 1 detail the update and retrieval procedure.

\begin{figure}[!htbp]
\centering\includegraphics[width=\textwidth]{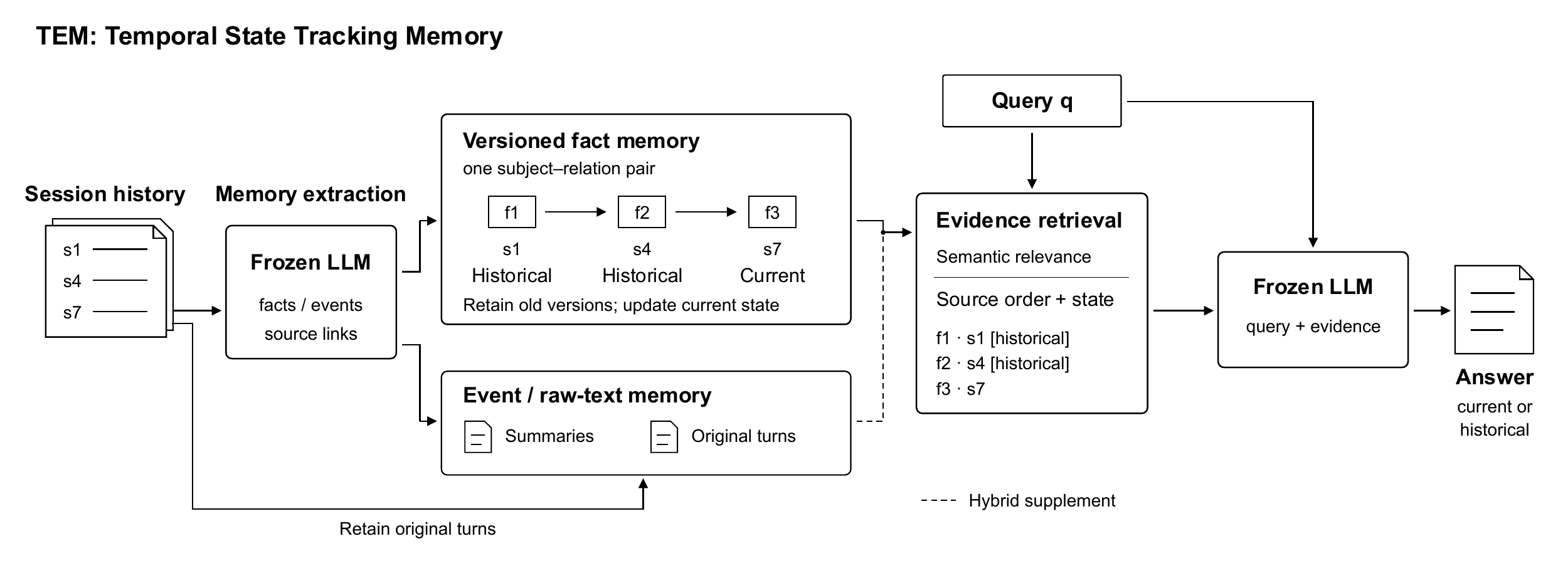}
\caption{TEM framework. Fact versions are linked to source sessions. Retrieved facts appear in source order, with historical labels on superseded versions. Dashed connections supply event and original-text evidence.}\label{fig:1}
\end{figure}

\subsection{Fact Versions with Provenance}

A fact record contains the subject, relation, object, source session and source index. Records with the same subject and relation belong to one version group:

\begin{equation}f=(s,r,o,\sigma,\tau),\qquad g(f)=(\operatorname{norm}(s),r)\label{eq:1}\end{equation}

The source session identifies the original record. Its index determines the position of that record in the version sequence. We group records by subject and relation, normalising subject names to lowercase. Excluding the object from the key lets different values of a relation remain in the same group.

An update changes a record's status but leaves its value and provenance intact. The rule assumes single-valued updates arriving in session order. Source indices record when information entered the memory; they do not establish when the underlying event took effect.

\subsection{History-Preserving Incremental Updates}

TEM appends facts from each new session to the existing version set without overwriting earlier values (Fig. 2a):

\begin{equation}F_t=F_{t-1}\cup\operatorname{AttachSource}(\Delta F_t,\sigma_t)\label{eq:2}\end{equation}

A frozen model extracts facts from the new session. TEM checks their format and consistency with the source text before attaching the session provenance.

TEM determines status from the complete version group before selecting evidence. A version is historical if its source index is smaller than the maximum index in that subject--relation group:

\begin{equation}\begin{aligned}\tau^*(g)&=\max\{\tau(f):f\in F_g\},\\h(f)&=\mathbf{1}[\tau(f)<\tau^*(g(f))].\end{aligned}\label{eq:3}\end{equation}

The historical attribute is one for a superseded version and zero when no later source exists. Both the value and its provenance remain in memory. A missed retrieval of the latest version therefore cannot make an older record appear current.

The label identifies a record as historical; its source index places it among the earlier versions. All records at the maximum source index are treated as current. This rule leaves conflicting values within the same session unresolved.

\subsection{Relevance Retrieval and Temporal Evidence}

Retrieval selects facts by relevance without penalising older versions for their age. TEM then places the selected facts in source order (Fig. 2b). The query, source order and version status are available to the generator when it chooses an answer.

Candidate scores combine similarity from the shared vector retriever with a weight for the relation type:

\begin{equation}u(f,q)=\cos(e(f),e(q))\bigl(0.4+0.6w(f)\bigr)\label{eq:4}\end{equation}

The relation weights are 1.0 for residence and employer, 0.9 for occupation, 0.8 for possession, and 0.6 for likes and dislikes. We apply this scoring rule to all versions and retrieve at most 15, taking those with the highest scores.

TEM sorts the selected records by ascending source index, keeping the fact content and source session. Superseded versions receive a [historical] label. The latest version has no label. Evidence construction and answering follow:

\begin{equation}\begin{aligned}B_q&=\operatorname{Render}(\operatorname{Sort}_{\tau}(\operatorname{TopK}(F,q)),\sigma,h),\\a&=\operatorname{LLM}(q,B_q).\end{aligned}\label{eq:5}\end{equation}

The retrieval budget counts fact records. Sorting and labelling leave their values unchanged. TEM does not use the question to exclude current or historical versions in advance; the generator chooses among the retrieved versions using the query and the supplied temporal information.

\subsection{Update and Evidence Construction Algorithm}

Algorithm 1 specifies version updates and retrieval, using frozen models for fact extraction and answer generation.

\begin{algorithm}[H]
\caption{Fact-version updates and temporal evidence construction}
\label{alg:tem}
\footnotesize
\textbf{Input:} stored versions $F$; new session $D$ and source $\sigma$; optional query $q$.\\
\textbf{Parameters:} candidate count $k$; relation weights $w$.\\
\textbf{Output:} updated $F$; evidence $B$ and answer $a$ if queried.
\begin{algorithmic}[1]
\State $\Delta F\gets\operatorname{ExtractAndValidate}(D,F)$
\State $F\gets F\cup\operatorname{AttachSource}(\Delta F,\sigma)$
\For{each subject--relation group $g\in\operatorname{Groups}(F)$}
  \State $\tau^*(g)\gets\max\{\tau(f):f\in F_g\}$
  \For{each version $f\in F_g$}
    \State $h(f)\gets\mathbf{1}[\tau(f)<\tau^*(g)]$
  \EndFor
\EndFor
\State \textbf{if} no query \textbf{then return} $F$
\State $C\gets\operatorname{TopK}(F,q,k)$, scored by Eq.~(4)
\State $C\gets\operatorname{Sort}_{\tau}(C)$, ascending source index
\State $B\gets\varnothing$
\For{each candidate $f\in C$}
  \State $b\gets\operatorname{RenderFactAndSource}(f)$
  \State \textbf{if} $h(f)=1$ \textbf{then} append \texttt{[historical]}
  \State $B\gets\operatorname{Concat}(B,b)$
\EndFor
\State $a\gets\operatorname{FrozenLLM}(q,B)$
\State \Return $F,B,a$
\end{algorithmic}
\end{algorithm}

AttachSource links each fact to its source session and index.

\begin{figure}[!htbp]
\centering\includegraphics[width=\textwidth]{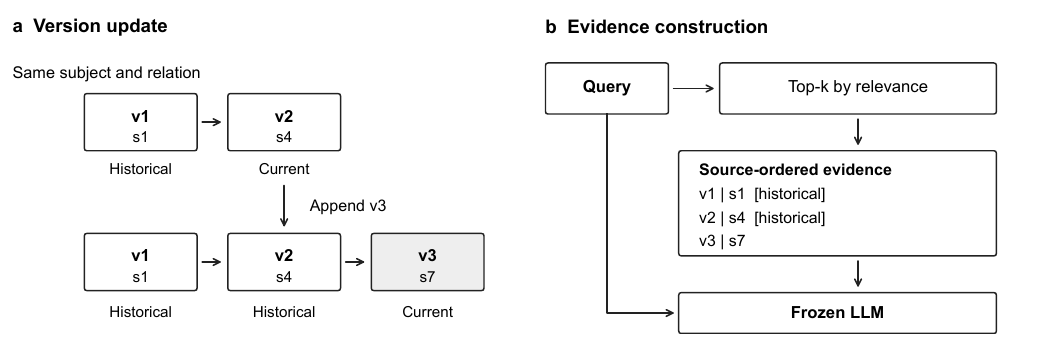}
\caption{Fact-version updates and evidence construction. a, Appending a new version updates the status of retained records. b, Retrieval selects relevant facts, which are then ordered by source and labelled if historical. Symbols v and s denote a fact version and its source session.}\label{fig:2}
\end{figure}

\subsection{Event Memory and Original-Text Evidence}

Some information does not fit a structured fact value, including event context and conditions on behaviour. TEM stores event summaries, free-text memories and original turns for this purpose. Source-session links connect these records to fact versions. Hybrid retrieval draws on both sources of evidence.

We use the fact-memory configuration TEM-Status for state-update experiments and TEM-Hybrid-State for public question answering. The hybrid configuration combines the evidence sources under a common budget. Its unmarked control uses the same candidates in the same order (Sections 4.1 and 4.6).

\section{Experiments and Results}

\subsection{Experimental Setup}

The state-update benchmark contains 60 personas, 600 sessions and 420 questions involving residence, occupation and drinks. There are 180 current-state questions, 180 other historical questions and 60 questions about residence immediately before the latest move. For a historical question, the reference answer is the most recent applicable fact before the target update. Running six memory configurations with four backbones produced 10,080 answers.

We compared TEM with naive RAG, session-aware RAG and full context. Session-aware RAG keeps retrieved source turns in session order. The fact-memory comparison uses three configurations. Overwrite keeps only the latest version. Unmarked keeps all versions in source order, and TEM-Status adds historical labels to the same content without changing that order.

Experiments used an A100 40 GB GPU. Extraction and answering within a configuration used the same frozen backbone. We retrieved at most 15 candidates with all-mpnet-base-v2 and set temperature to zero and the random seed to 42. Llama and Qwen ran in BF16 with a 512-token answer limit; Qwen thinking was disabled. For gpt-oss, we used MXFP4, low reasoning effort and a 4,096-token answer limit. All backbones had an 8,192-token extraction output limit.

We measured state-update performance by key-entity hit rate, counting an answer as a hit if it contained the reference entity under case-insensitive matching. Public question answering used strict semantic scoring, which required the answer to satisfy both factual and temporal conditions.

\subsection{Overall Question Answering}

With fact content and source order held constant, TEM-Status improved overall key-entity hit rates by 17.14--55.95 percentage points over Unmarked across the four backbones (Table 1).

\begin{table}[!htbp]
\caption{Effect of historical labels on overall key-entity hit rate}
\label{tab:1}
\centering\footnotesize
\setlength{\tabcolsep}{1.5pt}
\renewcommand{\arraystretch}{1.18}
\begin{tabular}{@{}>{\raggedright\arraybackslash}p{0.28\linewidth}>{\centering\arraybackslash}p{0.22\linewidth}>{\centering\arraybackslash}p{0.22\linewidth}>{\centering\arraybackslash}p{0.22\linewidth}@{}}
\toprule
\textbf{Backbone} & \textbf{Unmarked (\%)} & \textbf{TEM (\%)} & \textbf{Gain (pp)} \\
\midrule
Llama-3.1-8B & 19.76 & 75.71 & +55.95 \\
Qwen3.5-4B & 48.10 & 97.62 & +49.52 \\
Qwen3.5-9B & 74.52 & 91.67 & +17.14 \\
gpt-oss-20b & 55.24 & 96.67 & +41.43 \\
\bottomrule
\end{tabular}
\par\smallskip\noindent\begin{minipage}{0.90\textwidth}\footnotesize Each configuration contains 420 questions. Hit rates are percentages; gains are percentage points (pp).\end{minipage}
\end{table}

TEM-Status achieved overall key-entity hit rates of 75.71--97.62\%, exceeding session-aware RAG by 1.43--40.00 percentage points and full context by 2.14--30.24 points (Table 2).

\begin{table}[!htbp]
\caption{Overall key-entity hit rates across four backbones and six configurations (\%)}
\label{tab:2}
\centering\footnotesize
\setlength{\tabcolsep}{1.5pt}
\renewcommand{\arraystretch}{1.18}
\begin{tabular}{@{}>{\raggedright\arraybackslash}p{0.27\linewidth}>{\centering\arraybackslash}p{0.17\linewidth}>{\centering\arraybackslash}p{0.17\linewidth}>{\centering\arraybackslash}p{0.17\linewidth}>{\centering\arraybackslash}p{0.17\linewidth}@{}}
\toprule
\textbf{Method} & \textbf{Llama-3.1-8B} & \textbf{Qwen3.5-4B} & \textbf{Qwen3.5-9B} & \textbf{gpt-oss-20b} \\
\midrule
Naive RAG & 14.52 & 42.38 & 14.05 & 46.43 \\
Session-aware RAG & 56.67 & 57.62 & 79.76 & 95.24 \\
Full context & 49.76 & 95.48 & 87.38 & 66.43 \\
Overwrite & 43.33 & 29.76 & 41.43 & 34.76 \\
Retain, unmarked & 19.76 & 48.10 & 74.52 & 55.24 \\
TEM-Status & \textbf{75.71} & \textbf{97.62} & \textbf{91.67} & \textbf{96.67} \\
\bottomrule
\end{tabular}
\par\smallskip\noindent\begin{minipage}{0.90\textwidth}\footnotesize Each entry covers 420 questions. Bold indicates the highest value for a backbone.\end{minipage}
\end{table}

\subsection{Historical States after Repeated Updates}

Overwrite recovered few historical residence values, with hit rates of 0--8.33\%. Keeping earlier versions without labels raised the rates to 41.67--81.67\%. TEM-Status reached 96.67--100.00\% after historical labels were added to the same evidence (Fig. 3a).

TEM-Status exceeded session-aware RAG by 10.00--88.33 percentage points and full context by 3.33--40.00 points (Table 3).

\begin{figure}[!htbp]
\centering\includegraphics[width=\textwidth]{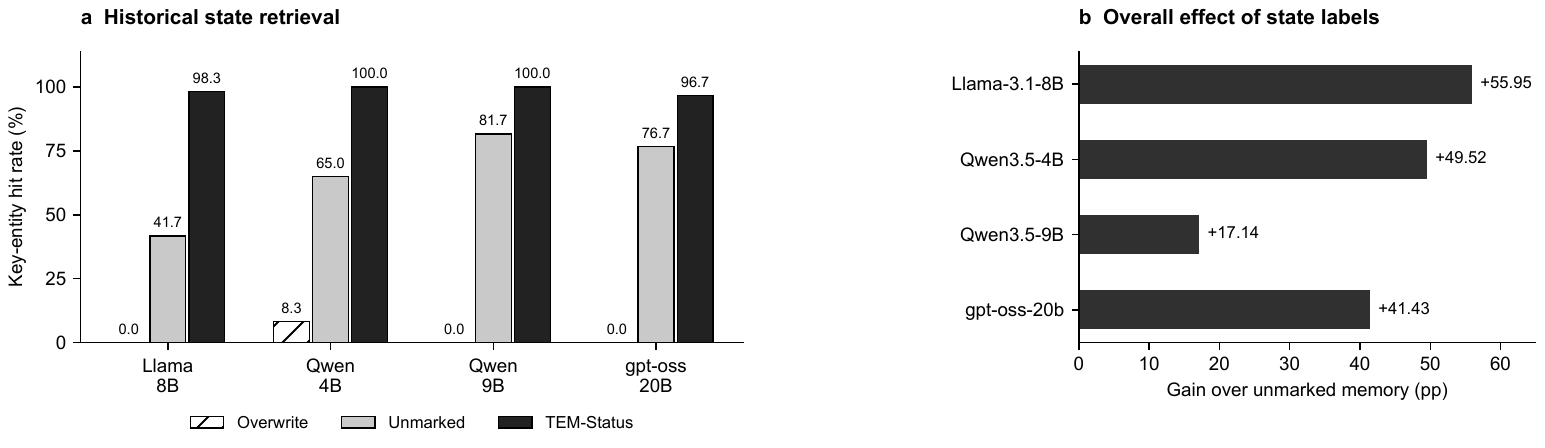}
\caption{Effects of version retention and historical labels. a, Historical-residence hit rates (60 questions). b, Overall gains over Unmarked (420 questions). pp denotes percentage points.}\label{fig:3}
\end{figure}

\begin{table}[!htbp]
\caption{Historical-residence retrieval after repeated updates (key-entity hit rate, \%)}
\label{tab:3}
\centering\footnotesize
\setlength{\tabcolsep}{1.5pt}
\renewcommand{\arraystretch}{1.18}
\begin{tabular}{@{}>{\raggedright\arraybackslash}p{0.27\linewidth}>{\centering\arraybackslash}p{0.17\linewidth}>{\centering\arraybackslash}p{0.17\linewidth}>{\centering\arraybackslash}p{0.17\linewidth}>{\centering\arraybackslash}p{0.17\linewidth}@{}}
\toprule
\textbf{Method} & \textbf{Llama-3.1-8B} & \textbf{Qwen3.5-4B} & \textbf{Qwen3.5-9B} & \textbf{gpt-oss-20b} \\
\midrule
Naive RAG & 0.00 & 0.00 & 0.00 & 5.00 \\
Session-aware RAG & 10.00 & 88.33 & 20.00 & 86.67 \\
Full context & 58.33 & 96.67 & 73.33 & 91.67 \\
Overwrite & 0.00 & 8.33 & 0.00 & 0.00 \\
Retain, unmarked & 41.67 & 65.00 & 81.67 & 76.67 \\
TEM-Status & \textbf{98.33} & \textbf{100.00} & \textbf{100.00} & \textbf{96.67} \\
\bottomrule
\end{tabular}
\par\smallskip\noindent\begin{minipage}{0.90\textwidth}\footnotesize Each entry covers 60 questions. Bold indicates the highest value for a backbone.\end{minipage}
\end{table}

\subsection{Ablation of State Labels}

Adding historical labels gave 235 more hits with Llama-3.1-8B, 208 with Qwen3.5-4B, 72 with Qwen3.5-9B and 174 with gpt-oss-20b over 420 questions (Table 4). These counts measure the net change in hits with fact content and source order held fixed.

\begin{table}[!htbp]
\caption{Overall hit counts with and without historical labels}
\label{tab:4}
\centering\footnotesize
\setlength{\tabcolsep}{1.5pt}
\renewcommand{\arraystretch}{1.18}
\begin{tabular}{@{}>{\raggedright\arraybackslash}p{0.28\linewidth}>{\centering\arraybackslash}p{0.22\linewidth}>{\centering\arraybackslash}p{0.22\linewidth}>{\centering\arraybackslash}p{0.22\linewidth}@{}}
\toprule
\textbf{Backbone} & \textbf{Unmarked hits} & \textbf{TEM hits} & \textbf{Net difference} \\
\midrule
Llama-3.1-8B & 83 & 318 & +235 \\
Qwen3.5-4B & 202 & 410 & +208 \\
Qwen3.5-9B & 313 & 385 & +72 \\
gpt-oss-20b & 232 & 406 & +174 \\
\bottomrule
\end{tabular}
\par\smallskip\noindent\begin{minipage}{0.90\textwidth}\footnotesize Each configuration contains 420 questions. Net gains are differences in total hit counts.\end{minipage}
\end{table}

The selected examples illustrate errors in identifying the time or relation requested by a question (Fig. 4). Under the control configurations, the same backbone sometimes used an old value as current, confused versions around an update or returned an unrelated relation. TEM recovered the requested value in each of these examples.

\begin{figure}[!htbp]
\centering\includegraphics[width=\textwidth]{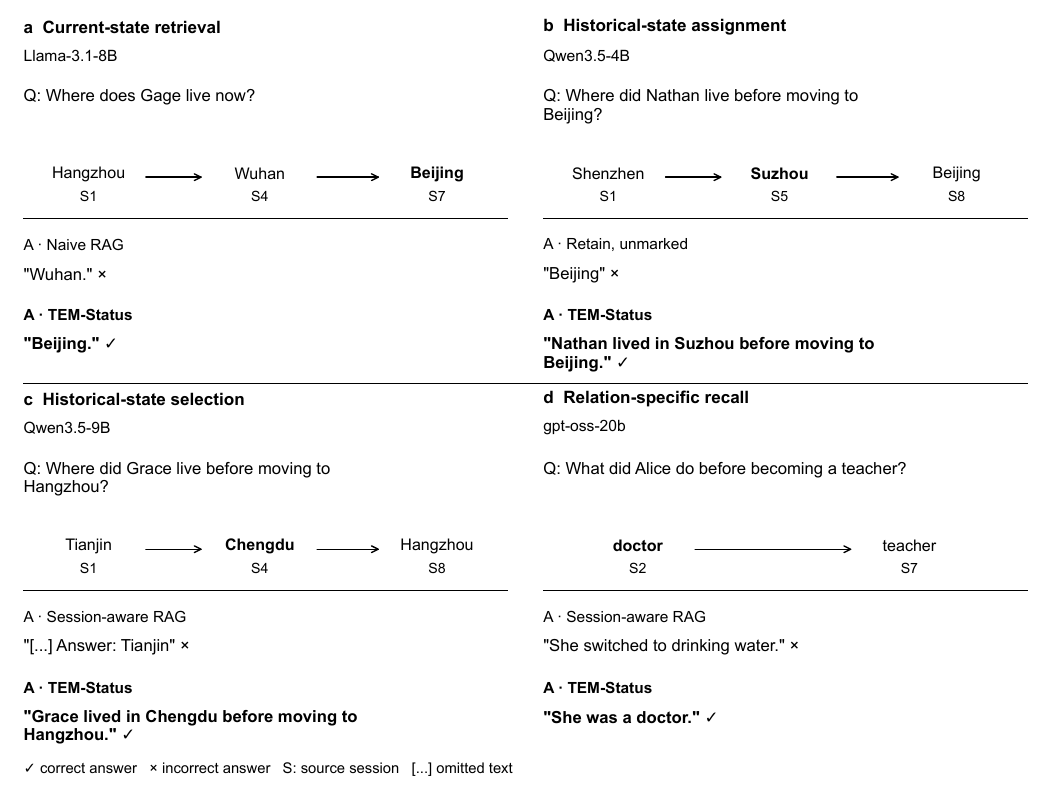}
\caption{Selected successful corrections across backbones. Arrows show source-session order; bold states give the reference values. Check marks and crosses indicate correct and incorrect answers. The control configuration is identified in each panel. Panel c marks omitted text with [\ldots{}].}\label{fig:4}
\end{figure}

Historical labels increased hit rates by 18.33--56.67 percentage points on the 60 residence questions (Table 5). The remaining 360 questions combine current-state and other historical queries. Gains of 16.94--55.83 points on this group show that the improvement extended beyond the historical-residence subset.

\begin{table}[!htbp]
\caption{Key-entity hit rates and label gains by query group}
\label{tab:5}
\centering\footnotesize
\setlength{\tabcolsep}{1.5pt}
\renewcommand{\arraystretch}{1.18}
\begin{tabular}{@{}>{\raggedright\arraybackslash}p{0.19\linewidth}>{\centering\arraybackslash}p{0.25\linewidth}>{\centering\arraybackslash}p{0.12\linewidth}>{\centering\arraybackslash}p{0.25\linewidth}>{\centering\arraybackslash}p{0.12\linewidth}@{}}
\toprule
\textbf{Backbone} & \textbf{Other queries\newline Unmarked / TEM} & \textbf{Gain (pp)} & \textbf{Pre-move residence\newline Unmarked / TEM} & \textbf{Gain (pp)} \\
\midrule
Llama-3.1-8B & 16.11 / 71.94 & +55.83 & 41.67 / 98.33 & +56.67 \\
Qwen3.5-4B & 45.28 / 97.22 & +51.94 & 65.00 / 100.00 & +35.00 \\
Qwen3.5-9B & 73.33 / 90.28 & +16.94 & 81.67 / 100.00 & +18.33 \\
gpt-oss-20b & 51.67 / 96.67 & +45.00 & 76.67 / 96.67 & +20.00 \\
\bottomrule
\end{tabular}
\par\smallskip\noindent\begin{minipage}{0.90\textwidth}\footnotesize Hit rates are percentages and gains are percentage points. Counts for other queries are obtained by subtracting the historical-residence subset from the overall counts.\end{minipage}
\end{table}

\subsection{Answer-Evidence Length}

TEM used 94.05--117.55 tokens of answer evidence on average, 48.75--56.73\% less than session-aware RAG (Table 6). These lengths exclude memory construction and embedding computation.

\begin{figure}[!htbp]
\centering\includegraphics[width=\textwidth]{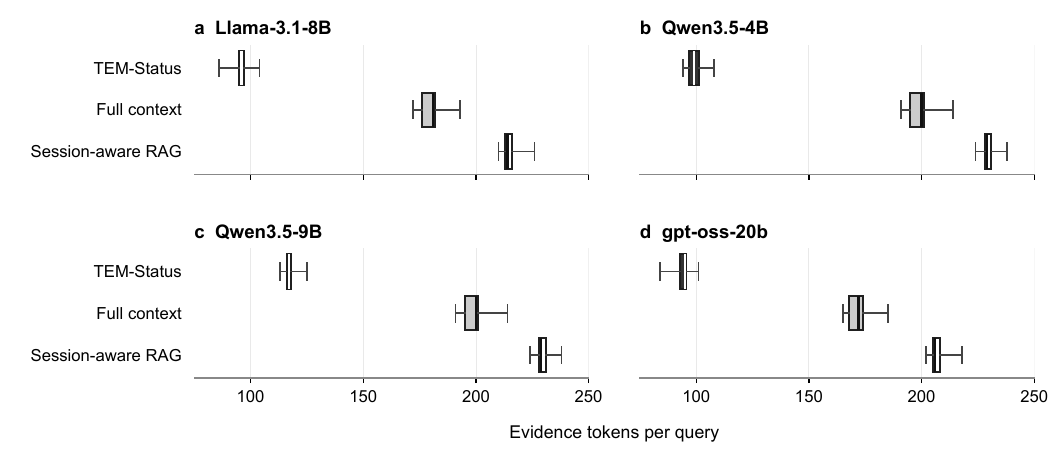}
\caption{Answer-evidence length distributions, with 420 queries per configuration. Boxes show the interquartile range, centre lines the median, and whiskers the minimum and maximum.}\label{fig:5}
\end{figure}

\begin{table}[!htbp]
\caption{Mean evidence length during answering}
\label{tab:6}
\centering\footnotesize
\setlength{\tabcolsep}{1.5pt}
\renewcommand{\arraystretch}{1.18}
\begin{tabular}{@{}>{\raggedright\arraybackslash}p{0.23\linewidth}>{\centering\arraybackslash}p{0.22\linewidth}>{\centering\arraybackslash}p{0.18\linewidth}>{\centering\arraybackslash}p{0.18\linewidth}>{\centering\arraybackslash}p{0.13\linewidth}@{}}
\toprule
\textbf{Backbone} & \textbf{Session-aware RAG} & \textbf{Full context} & \textbf{TEM-Status} & \textbf{Saving (\%)} \\
\midrule
Llama-3.1-8B & 214.72 & 180.13 & 95.93 & 55.32 \\
Qwen3.5-4B & 229.36 & 199.52 & 99.25 & 56.73 \\
Qwen3.5-9B & 229.36 & 199.52 & 117.55 & 48.75 \\
gpt-oss-20b & 206.55 & 171.95 & 94.05 & 54.47 \\
\bottomrule
\end{tabular}
\par\smallskip\noindent\begin{minipage}{0.90\textwidth}\footnotesize Evidence length is measured in tokens per question. Savings are relative to session-aware RAG.\end{minipage}
\end{table}

TEM evidence ranged from 84 to 125 tokens across all queries (Fig. 5). Even the shortest inputs from session-aware RAG and full context were longer, at 202 and 165 tokens.

\subsection{Public Multi-Session Question Answering}

We evaluated public question answering on 10 LoCoMo conversations containing 1,540 questions: 282 multi-hop, 321 temporal, 96 open-domain and 841 single-hop. Llama-3.1-8B and Qwen3.5-9B each ran four configurations, producing 12,320 answers in total. TEM-Hybrid-State, TEM-Hybrid-Unmarked and session-aware RAG shared a 1,024-token evidence limit. The two hybrid configurations used identical candidates and ordering. A long-context configuration provided a further reference. All memories were built from the source conversations only.

For each backbone, we kept the answer prompt, zero temperature, seed 42 and 1,024-token answer limit fixed across configurations. Qwen thinking was disabled. Inputs to the long-context reference were truncated at the 32,768-token model window; this affected 1,151 questions per backbone. We used gpt-oss-20b to score factual and temporal correctness against the reference answers and supporting turns.

Ambiguity in the reference information left 216 answers unresolved. We retained the full question count as the denominator and assigned these cases zero or one to obtain score bounds. Eleven truncated or empty answers counted as incorrect.

Under the same retrieval budget, Llama scored 27.53--29.22\% with TEM and 17.92--19.61\% with session-aware RAG (Table 7). Qwen scored 35.84--37.53\% and 23.70--25.26\%, respectively. Even subtracting the RAG upper bound from the TEM lower bound left gaps of 7.92 and 10.58 percentage points.

\begin{table}[!htbp]
\caption{Strict semantic scores on LoCoMo for two backbones (\%)}
\label{tab:7}
\centering\footnotesize
\setlength{\tabcolsep}{1.5pt}
\renewcommand{\arraystretch}{1.18}
\begin{tabular}{@{}>{\raggedright\arraybackslash}p{0.115\linewidth}>{\raggedright\arraybackslash}p{0.215\linewidth}>{\centering\arraybackslash}p{0.12\linewidth}>{\centering\arraybackslash}p{0.12\linewidth}>{\centering\arraybackslash}p{0.12\linewidth}>{\centering\arraybackslash}p{0.12\linewidth}>{\centering\arraybackslash}p{0.12\linewidth}@{}}
\toprule
\textbf{Backbone} & \textbf{Memory} & \textbf{Overall} & \textbf{Temporal} & \textbf{Multi-hop} & \textbf{Single-hop} & \textbf{Open-domain} \\
\midrule
Llama-3.1-8B & TEM-Hybrid-State & 27.53--29.22 & 14.02--17.76 & 17.73--18.09 & 37.57--38.88 & 13.54--15.62 \\
Llama-3.1-8B & TEM-Hybrid-Unmarked & 27.40--28.96 & 14.33--17.45 & 14.89--15.25 & 38.17--39.48 & 13.54--15.62 \\
Llama-3.1-8B & Session-aware RAG & 17.92--19.61 & 8.72--11.84 & 12.06--12.77 & 23.90--25.33 & 13.54--15.62 \\
Llama-3.1-8B & Long-context ref. & 37.66--39.55 & 14.64--17.13 & 24.11--24.82 & 53.51--55.65 & 15.62--16.67 \\
Qwen3.5-9B & TEM-Hybrid-State & 35.84--37.53 & 38.63--42.06 & 22.70--23.05 & 41.38--42.81 & 16.67--18.75 \\
Qwen3.5-9B & TEM-Hybrid-Unmarked & 35.97--37.66 & 38.32--41.43 & 25.18--25.89 & 40.78--42.21 & 17.71--19.79 \\
Qwen3.5-9B & Session-aware RAG & 23.70--25.26 & 23.99--27.10 & 19.86--20.57 & 25.92--27.23 & 14.58--15.62 \\
Qwen3.5-9B & Long-context ref. & 53.90--56.17 & 48.91--52.96 & 28.72--30.14 & 66.94--68.85 & 30.21--32.29 \\
\bottomrule
\end{tabular}
\par\smallskip\noindent\begin{minipage}{0.90\textwidth}\footnotesize Each configuration contains 1,540 questions. Ranges are bounds obtained by assigning unresolved scores zero or one, not confidence intervals. The first three configurations have a 1,024-token evidence limit; the long-context reference has a 32,768-token window.\end{minipage}
\end{table}

Historical labels made little difference to overall LoCoMo scores. Mean TEM evidence lengths were 1,014.6 tokens for Llama and 1,017.5 for Qwen, close to session-aware RAG. The long-context reference scored higher with much larger inputs: 29,671.8 and 30,412.9 tokens, respectively. These public questions require event details and multi-hop associations as well as state information. The similar scores of the marked and unmarked variants suggest that hybrid evidence organisation accounts for most of the observed gain over RAG.

\subsection{Results Reported for Existing Memory Systems}

Published memory-system evaluations provide further context for these results \cite{ref10,ref11,ref12,ref13}. Their generators and scoring metrics are recorded alongside the reported scores in Table 8. TEM was evaluated with open-weight generators and strict semantic scoring.

\begin{table}[!htbp]
\caption{Published results and evaluation settings of representative memory methods}
\label{tab:8}
\centering\footnotesize
\setlength{\tabcolsep}{1.5pt}
\renewcommand{\arraystretch}{1.18}
\begin{tabular}{@{}>{\raggedright\arraybackslash}p{0.15\linewidth}>{\raggedright\arraybackslash}p{0.16\linewidth}>{\centering\arraybackslash}p{0.16\linewidth}>{\centering\arraybackslash}p{0.1\linewidth}>{\centering\arraybackslash}p{0.12\linewidth}>{\centering\arraybackslash}p{0.12\linewidth}>{\centering\arraybackslash}p{0.12\linewidth}@{}}
\toprule
\textbf{Method} & \textbf{Generator} & \textbf{Benchmark} & \textbf{Metric} & \textbf{Overall} & \textbf{Temporal} & \textbf{Single-hop} \\
\midrule
A-MEM \cite{ref10} & GPT-4o-mini & LoCoMo & F1 & --- & 45.85 & 44.65 \\
MemoryOS \cite{ref12} & GPT-4o-mini & LoCoMo & F1 & --- & 20.02 & 35.27 \\
Mem0 \cite{ref11} & GPT-4o-mini & LoCoMo & J & 66.88$\pm$0.15 & 55.51$\pm$0.34 & 67.13$\pm$0.65 \\
Mem0-graph \cite{ref11} & GPT-4o-mini & LoCoMo & J & 68.44$\pm$0.17 & 58.13$\pm$0.44 & 65.71$\pm$0.45 \\
Zep \cite{ref13} & GPT-4o-mini & LongMemEval-s & GPT-4o judge & 63.80 & 54.10 & --- \\
Zep \cite{ref13} & GPT-4o & LongMemEval-s & GPT-4o judge & 71.20 & 62.40 & --- \\
TEM-Hybrid & Llama-3.1-8B & LoCoMo & Strict judge & 27.53--29.22 & 14.02--17.76 & 37.57--38.88 \\
TEM-Hybrid & Qwen3.5-9B & LoCoMo & Strict judge & 35.84--37.53 & 38.63--42.06 & 41.38--42.81 \\
\bottomrule
\end{tabular}
\par\smallskip\noindent\begin{minipage}{0.90\textwidth}\footnotesize Scores are percentages. Generators, scoring rules and input budgets differ, so values are not a common ranking. A-MEM values are from v11, Table 1; MemoryOS values are from Table 2 of the published version. Mem0 reports standard deviations over 10 evaluations. TEM ranges bound unresolved scores. A dash indicates an unreported or inapplicable value.\end{minipage}
\end{table}

\FloatBarrier

\section{Discussion}

The state-update results separate two requirements of historical question answering. A memory must first keep earlier values accessible. Once those values are available, the generator still needs to identify which one applies. In the ablation, historical labels improved hit rates without changing the facts or their source order. Retrieval therefore did not account for the difference between the two configurations. The result supports treating version status as part of the evidence presented to the generator, even when it can in principle be inferred from the ordered records.

On LoCoMo, marked and unmarked hybrid memory obtained similar scores, and both exceeded session-aware RAG under the same evidence budget. The contrast with the state-update task helps define the role of the labels. They directly indicate whether a retained value has been superseded, but do not supply missing event details or cross-session associations. These are also needed for public dialogue questions. The hybrid-memory gain on LoCoMo is therefore better understood at the level of evidence organisation, without attributing it to historical labels alone.

TEM uses source-session order to assign state, assuming explicit single-valued updates. This is sufficient for the controlled update setting considered here. A delayed report may place an earlier event in a later session, and contradictory values can occur within one session. Preferences may also remain valid simultaneously. Handling these cases requires event-validity times or explicit treatment of conflicting relations beyond the present ordering rule.

\section{Conclusion}

TEM couples history-preserving fact storage with evidence that makes version status explicit. The update rule operates on complete version groups, and retrieval passes source order and historical labels to the generator. Across four backbones, retaining and labelling versions improved historical information use in state-update tasks with compact fact evidence. Hybrid memory also exceeded session-aware RAG on public multi-session questions at the same evidence budget. Together, the results support managing fact versions throughout storage and answering when information changes across conversations.

\section*{Statements and Declarations}

\subsection*{Funding}
This research was supported by Zhejiang Provincial Natural Science Foundation of China under Grant No. LQN26F020016.

\subsection*{Competing Interests}
The authors declare that they have no competing interests.

\subsection*{Data and Code Availability}

The public LoCoMo dataset used in this study is available from its official source \cite{ref24}. Illustrative code for the TEM mechanism and an overview of the main results are available at \url{https://github.com/sungkites/TEM}.

\bibliography{references}

\end{document}